\documentclass[11pt,a4paper]{article}
\usepackage[hyperref]{acl2019}
\usepackage{times}
\usepackage{latexsym}

\usepackage{url}

\usepackage{todonotes}
\usepackage{amssymb}
\usepackage{amsmath}
\usepackage{booktabs}
\usepackage{multirow}
\usepackage{algorithm}
\usepackage[noend]{algpseudocode}
\makeatletter
\let\OldStatex\Statex
\renewcommand{\Statex}[1][3]{%
  \setlength\@tempdima{\algorithmicindent}%
  \OldStatex\hskip\dimexpr#1\@tempdima\relax}
\makeatother
\algnewcommand{\LineComment}[1]{\State \(\triangleright\) #1}

\aclfinalcopy 

\title{Multimodal and Multi-view Models for Emotion Recognition}

\author{Gustavo Aguilar$^{\ddagger}$, Viktor Rozgi\'c$^\dagger$, Weiran Wang$^\dagger$  \and Chao Wang$^\dagger$ \\
	University of Houston$^\ddagger$ \\
	Amazon.com$^\dagger$ \\
  {\tt gaguilaralas@uh.edu}, {\tt \{rozgicv, weiranw, wngcha\}@amazon.com}\\}

\date{}

\begin{document}
\maketitle

\begin{abstract}
    Studies on emotion recognition (ER) show that combining lexical and acoustic information results in more robust and accurate models. The majority of the studies focus on settings where both modalities are available in training and evaluation. However, in practice, this is not always the case; getting ASR output may represent a bottleneck in a deployment pipeline due to computational complexity or privacy-related constraints. 
    To address this challenge, we study the problem of efficiently combining acoustic and lexical modalities during training while still providing a deployable acoustic model that does not require lexical inputs. 
    We first experiment with multimodal models and two attention mechanisms to assess the extent of the benefits that lexical information can provide. 
    Then, we frame the task as a multi-view learning problem to induce semantic information from a multimodal model into our acoustic-only network using a contrastive loss function. 
    Our multimodal model outperforms the previous state of the art on the USC-IEMOCAP dataset reported on lexical and acoustic information. Additionally, our multi-view-trained acoustic network significantly surpasses models that have been exclusively trained with acoustic features.
\end{abstract}

\section{Introduction}

The task of emotion recognition (ER) requires understanding the way humans interact to express their emotional state during conversations. Among others, emotions are encoded in both lexical and acoustic information where each modality contributes to the overall emotional state of a given speaker. However, in some situations, one modality can be more insightful to derive emotions than the other. For instance, the phrase \textit{``yeah... of course''} does not have enough lexical information to derive the right emotion, and it may all depend on the acoustic patterns. On the other hand, the phrase \textit{``I really miss my dog!''} does not need acoustic information to detect that the most likely emotion is sadness. Thus, recognizing emotions is not a trivial task because an emotional state can be easily shaped by many factors: context, word content, spectral and prosodic information, among others \citep{DBLP:journals/speech/BarbulescuRB17}. 

In this paper, we study the emotion recognition problem from the speech and language perspectives. We formally look into acoustic and lexical modalities with the aim of improving models that only use acoustic information. 
In the first part of this work, our goal is to assess the extent to which lexical information benefits acoustic models. 
We propose a multimodal method that is inspired by the way humans process emotions in a conversation. That is, lexical and acoustic information is simultaneously perceived at every word step. Hence, we introduce the concept of acoustic words: word-level representations derived from acoustic features in a speech fragment. The acoustic word representations enable a natural combination of the modalities where lexical and acoustic features are aligned at the word level. Additionally, we leverage these representations with two attention mechanisms: modality-based and context-based attentions. The former mechanism prioritizes one of the modalities at each word step, whereas the latter mechanism focuses on the most important word representations across the entire utterance. Our multimodal approach outperforms the current state of the art on the USC-IEMOCAP dataset reported on lexical and acoustic modalities.

In the second part of this work, our goal is to induce semantic information from the proposed multimodal model into an acoustic model. 
We study a more challenging scenario where we establish that lexical information is available during training but not during the evaluation phase. Such restriction is commonly found in real-world applications, where transcripts or ASR outputs represent a bottleneck in a deployment pipeline due to computational complexity or privacy-related constraints. To address this challenge, we frame this task as a multi-view learning problem \citep{Blum:1998:CLU:279943.279962}. 
We induce lexical information from our multimodal model into the acoustic network during training while still providing a lexical-independent acoustic model for testing or deployment. That is, our acoustic model learns to capture semantic and contextual information without relying on explicit lexical inputs such as ASR or transcripts. This multi-view acoustic network significantly outperforms models that have been exclusively trained on acoustic features.

\section{Related Work}

Recognizing emotions is a complex task because it involves several ambiguous human interactions such as facial expressions, change in pitch or tone of voice, linguistic semantics and meaning, among others \citep{cowie2009perceiving, Provostetal2009}.
Many researchers have approached these challenges by extracting features from visual, acoustic, and lexical information. Early approaches rely on a variation of support vector machine (SVM) classifiers to learn emotional categories such as happiness, sadness, anger, and others \citep{rozgic2012ensemble, perezrosas-mihalcea-morency:2013:ACL2013, jinetal2015}. For instance, \citet{rozgic2012ensemble} use an automatically generated ensemble of trees whose nodes contain binary SVM classifiers for each emotional category. \citet{jinetal2015} also use multimodality, and their study focuses on comparing early and late-fusion methods. Consistently, researchers have found that multimodal approaches outperform unimodal ones.

Recent work has focused on different ways to fuse the acoustic, lexical, and visual modalities. However, we narrow the discussion to the acoustic and lexical modalities to align with the scope of the paper. In most of the cases, researchers have used concatenation to fuse the lexical and acoustic representations at different stages of their models. Other works have proposed multimodal pooling fusion \citep{Aldeneh:2017:PAL:3136755.3136760}, tensor fusion networks \citep{DBLP:journals/corr/ZadehCPCM17}, modality hierarchical fusion \citep{Majumderetal2018}, context-aware fusion with attention \citep{DBLP:journals/corr/abs-1803-07427}, and conversational memory networks (CMN) \cite{hazarika-EtAl:2018:N18-1}. Nevertheless, all the previous fusion techniques have been made at the utterance level, whereas our work focuses on multimodal fusion at the word level by introducing acoustic word representations. 
We compare our work to \citet{DBLP:journals/corr/abs-1803-07427} because they document the current best performance on lexical and acoustic information on the IEMOCAP dataset using the standard 10-fold speaker-exclusive cross-validation setting.

Closely related work on acoustic word embeddings has been made by \citet{DBLP:journals/corr/HeWL16}. They induce acoustic information into lexical representations at the character level in a multi-view unsupervised setting. We introduce the concept of acoustic word representations in a different way: we learn vector representations of words out of frame-level acoustic features. This allows us to align lexical and acoustic information at the word level, which simulates the way humans perceive emotions in conversations (i.e., both modalities are simultaneously perceived). 

We also explore multi-view settings to overcome the absence of lexical inputs during evaluation \citep{Blum:1998:CLU:279943.279962}. There are multiple options to conduct the experiments in this scenario \citep{DBLP:journals/corr/abs-1304-5634, DBLP:conf/icml/WangALB15}, such as deep cannonical correlation analysis (DCCA) \citep{Andrew:2013:DCC:3042817.3043076} and siamese networks with contrastive loss functions \citep{DBLP:journals/corr/HeWL16}. We use the latter approach in our experiments. To the best of our knowledge, there is no prior work trying to overcome the absence of lexical inputs by inducing lexical information into an acoustic model for the task of emotion recognition. 



\section{Methodology}

We describe the data representation and introduce the idea of acoustic words in Sections \ref{sec:input_representation} and \ref{sec:acoustic_words}. Then, we use this concept to define the multimodal architecture in Section \ref{sec:multimodal_model}. Finally, we explain the multi-view learning setting using the proposed multimodal model and our acoustic model in Section \ref{sec:multimodal_model}.


\subsection{Data Representation}
\label{sec:input_representation}

\noindent\textbf{Acoustic features}. We extract frame-level features using OpenSMILE\footnote{\url{audeering.com/technology/opensmile/}} \citep{Eyben:2013:RDO:2502081.2502224}. We use the Computational Paralinguistic Challenge (ComParE) feature-set introduced by \citet{schuller2013interspeech} for the InterSpeech emotion recognition challenge. These features include energy, spectral, MFCC, and other low-level descriptors. The InterSpeech ComParE 2013 features are fairly standard and well-documented. Additionally, we normalize these features using z-standardization before feeding them into our models.

\noindent\textbf{Lexical features}. We use word embeddings to represent the lexical information. Specifically, we employ deep contextualized word representations using the language model ELMo 
\citep{peters-EtAl:2018:N18-1}. ELMo represents words as vectors that are entirely built out of characters. 
This allows us to overcome the problem of out-of-vocabulary words by always having a vector based on morphological clues for any given word. Additionally, these representations have proven to capture syntax and semantics aspects as well as the diversity of the linguistic context of words (e.g., polysemy).

\subsection{Acoustic Words}
\label{sec:acoustic_words}


Previous studies usually extract features from the modalities in independent modules, and then they concatenate the corresponding utterance representations from the acoustics and lexical features to feed into the next layers of their models. However, we argue that a more natural way to understand emotions is to align lexical and acoustic information, which simulates the way humans process both modalities simultaneously. Thus, we introduce the concept of acoustic word representations (see Figure \ref{fig:multimodal_model}). 
These representations are extracted from frame-level features by taking the output of a bidirectional LSTM at every segment. Note that this procedure requires the word alignment information. 
Additionally, we exclude frames that do not belong to the words of the speaker. 
This reduces any potential bias towards other people's emotional states as well as environmental noise.

\subsection{Hierarchical Multimodal Model}
\label{sec:multimodal_model}

\begin{figure}[t!]
	\centering
	\includegraphics[width=0.8\linewidth]{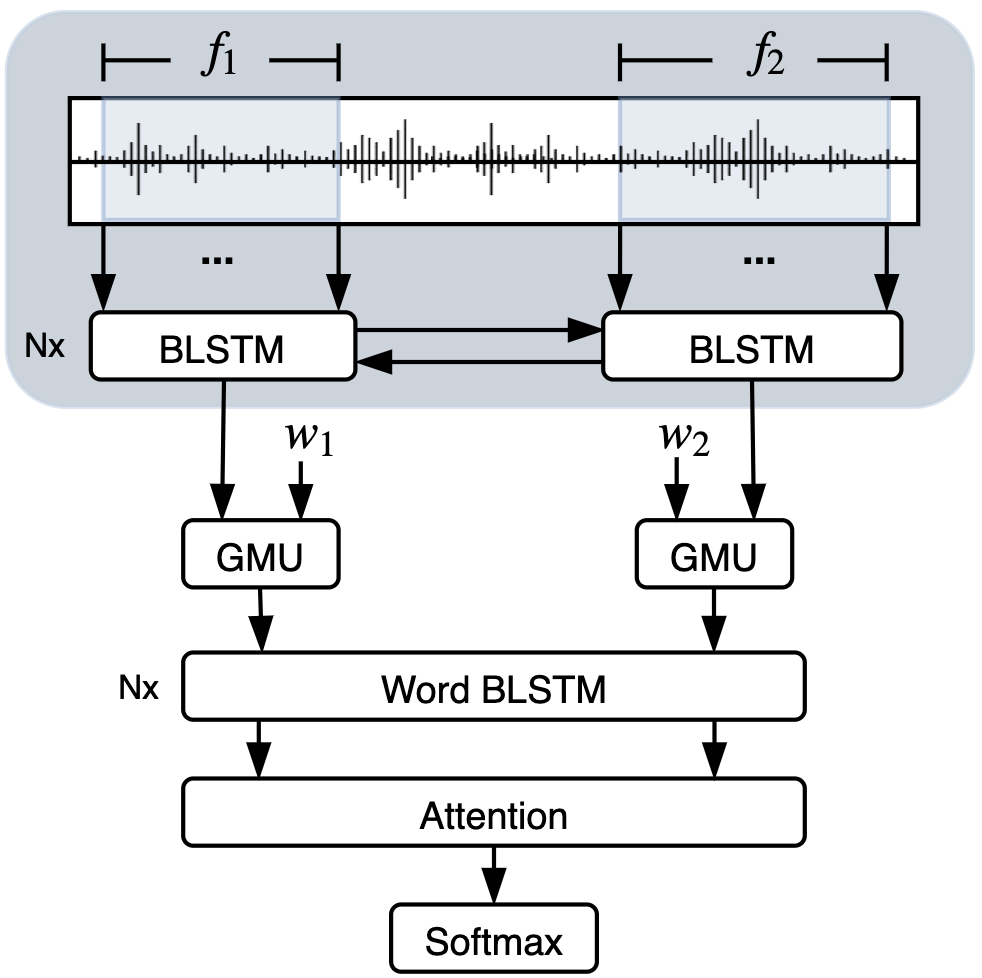}
	\caption{ The multimodal model. The shadowed box incloses the acoustic word mechanism, whose output is fed into the GMU unit along with the lexical word representation at each timestep. The model can have N layers of BLSTM at the frame and word levels. }
	\label{fig:multimodal_model}
\end{figure}

Our goal is to provide a neural network model that efficiently combines acoustic and lexical information for emotion recognition. We propose a hierarchical multimodal model that uses: 1) acoustic word representations derived from frame-level features, 2) a modality-based attention mechanism at the word level that prioritizes one modality over the other, and 3) a context-based attention mechanism that emphasizes the most relevant parts in the entire utterance. 
In Figure \ref{fig:multimodal_model}, the shadowed box represents the low level of the hierarchy, where the frame features are used to generate the acoustic word representation. The high level of the model is where the word representations from each modality are combined. 

\noindent\textbf{Modality-based attention}. The idea of the modality-based attention is to prioritize one of the modalities at the word level. That is, when the lexical features are more relevant to capture emotions (i.e., informative words are used), the model should prioritize such features and vice versa (i.e., arousal and pitch levels increase). To achieve this behavior, we incorporate the bimodal version of the GMU cell proposed by \citet{arevalo2017gated}. The GMU equations are as follows:
\begin{equation} \label{eq:attention-modality}
	\begin{split}
		h_a =& ~\mathrm{tanh}(\mathrm{W}_a x_a + b_a) \\
		h_l =& ~\mathrm{tanh}(\mathrm{W}_l x_l + b_l) \\
		z   =&~ \sigma(\mathrm{W}_z [x_a, x_l] + b_z) \\
		h   =&~ z * h_a + (1 - z) * h_l 
	\end{split}
\end{equation}

\noindent where $x_a$ and $x_l$ are the acoustic and lexical input vectors, respectively. These inputs are concatenated ($[x_a, x_l]$) and then multiplied by $\mathrm{W}_z$ so that the concatenation can be projected into the same space of the hidden vectors $h_a$ and $h_l$. Finally, $z$ is multiplied by the hidden acoustic vector $h_a$, and $(1- z)$ by the hidden lexical vector $h_l$. By adding the result of these products, the model incorporates a complementary mechanism over the modalities, which allows prioritizing one over the other when necessary.

\begin{figure*}[t!]
	\centering
	\includegraphics[width=0.9\linewidth]{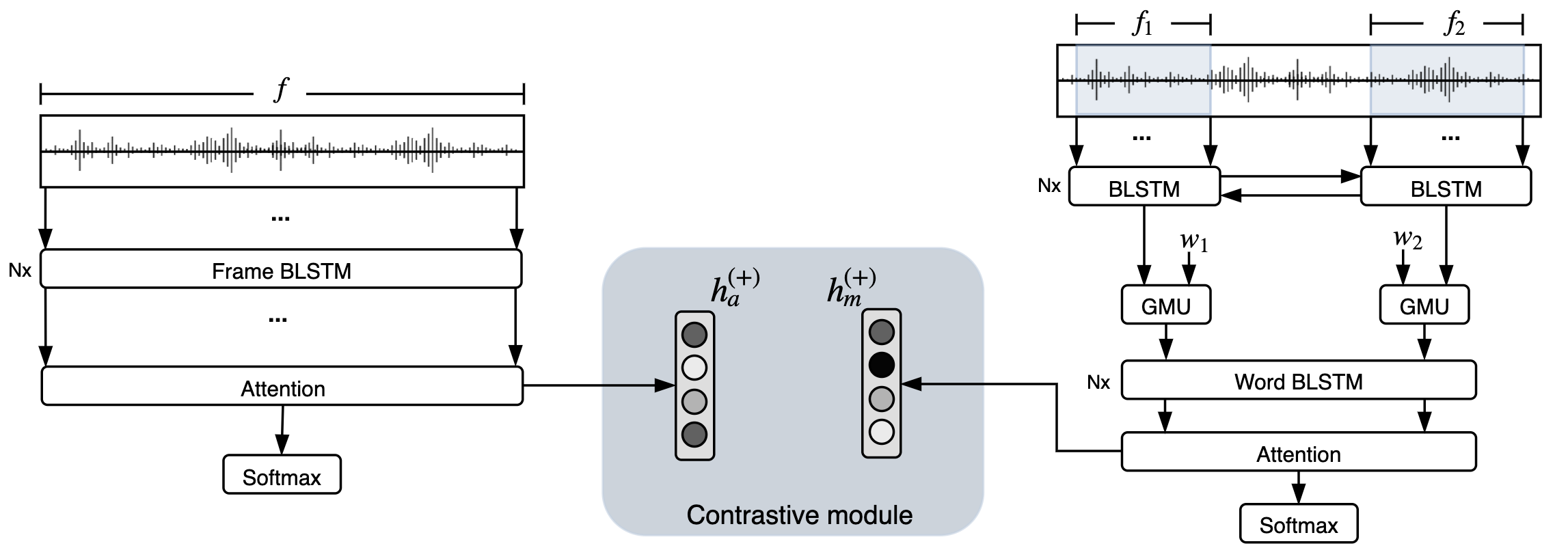}
	\caption{ The multi-view models. The view on the left is the acoustic model, and the view on the right is the multimodal model. The shadowed box in the middle is the contrastive loss module.}
	\label{fig:multiview_model}
\end{figure*}

\noindent\textbf{Context-based attention}. We use a fairly standard attention mechanism over the entire utterance that was introduced by \citet{DBLP:journals/corr/BahdanauCB14}. The idea is to concentrate mass probability over the words that capture emotional states along the sequence. Our attention mechanism uses the following equations:
\begin{equation} \label{eq:attention-context}
	\begin{split}
		e_i =&~ v^\intercal ~\mathrm{tanh}(\mathrm{W}_h h_i + b_h) \notag\\
		a_i =&~ \frac{\mathrm{exp}(e_i)}{\sum^{N}_{j=1}\mathrm{exp}(e_j)}, ~~~\text{where}~\sum_{i=1}^N a_i = 1 \notag\\
		z   =&~ \sum_{i=1}^N a_i h_i \notag
	\end{split}
\end{equation}
where $\mathrm{W}_h \in \mathbb{R}^{d_a \times d_h}$ and $b_h \in \mathbb{R}^{d_h}$ are trainable parameters of the model. The vector $v \in \mathbb{R}^{d_a}$ is the attention vector to be learned. Also, $d_a$ and $d_h$ are the dimensions of the attention layer and the hidden state, respectively. Then, we multiply the scalars $a_i$ and their corresponding hidden vectors $h_i$ to obtain our weighted sequence. The sum of the weighted vectors, $z$, is fed into a softmax layer.

\subsection{Multi-view Learning}
\label{sec:multiview_model}

A more realistic and challenging scenario happens when lexical information is not available during testing. In this case, our goal is to build an acoustic model that is capable of inferring some notion of semantic and contextual features by taking advantage of lexical information only available during training. To achieve this, we frame the problem as a multi-view learning task, where two disjoint networks share their learned information through the loss function \citep{Lian:2018:SER:3267935.3267946}. The fact that they are disjoint networks allows them to function without each other during evaluation.

Consider the acoustic and multimodal views $V_a$ and $V_m$. 
The acoustic view, $V_a$, is comprised of N layers of bidirectional LSTMs followed by an attention and a softmax layers. The multimodal view, $V_m$, follows the architecture described in Section \ref{sec:multimodal_model}.
As shown in Figure \ref{fig:multiview_model}, the view on the left, $V_a$, takes only the raw frame vectors, whereas the view on the right, $V_m$, takes the aligned frame and word vectors as inputs.
Each view learns an utterance representation of the emotions, $h_a$ and $h_m$, which are the outputs of their corresponding attention layers, as defined in Eq. \ref{eq:attention-context}. Since these vectors come from the same source of information (i.e., same speaker utterance), we assume that their emotion representations are similar. In general, we want vectors with similar emotions to be close and dissimilar ones to be far regardless of the modalities they use. To achieve this, we use the following contrastive loss function:

\noindent
\resizebox{\hsize}{!}{
  \begin{minipage}{1.2\hsize}
    \begin{align}\label{eq:contrastive_loss}
        \mathcal{L}_{c} = &\frac{1}{2N}\sum_i^N{\mathrm{max}(0, m + dis(h_{a_i}, h^+_{m_i}) - dis(h_{a_i}, h^-_{m_i}))} \notag \\
                          + &\frac{1}{2N}\sum_i^N{\mathrm{max}(0, m + dis(h_{m_i}, h^+_{a_i}) - dis(h_{m_i}, h^-_{a_i}))}
    \end{align}
    \vspace{0.2\baselineskip}
  \end{minipage}
}
where the $+$ and $-$ superscripts refer to positive (i.e., close) and negative (i.e., far) vectors. We force a margin of at least $m$ to keep negative samples separated from positive samples. We define $dis(v,w) = 1 - cos(v, w)$ as the function that calculates the distance between two vectors. Note that we determine cross-view pairs when comparing vectors because we want the models to induce similar information from different modalities. 
Additionally, choosing the negative samples can dramatically affect the performance of the models. 
For instance, for random samples that may not share acoustic or lexical properties, the models can easily satisfy the margin $m$ without forcing much learning. Instead, we want the models to find the nuances in acoustically similar samples that have different emotion labels.
Thus, besides random sampling, we also consider similar acoustic properties (e.g., valence, arousal, or dominance) that overlap among the emotions. 

In addition to the contrastive objective function, we use cross-entropy loss functions for the acoustic and multimodal views:
\begin{align} 
	\mathcal{L}_{a} =& - \frac{1}{N}\beta_a\sum_i^N y_i log(\hat{y}_i) \label{eq:aco-cross-entropy} \\
	\mathcal{L}_{m} =& - \frac{1}{N}\beta_m\sum_i^N y_i log(\hat{y}_i) \label{eq:lex-cross-entropy}
\end{align}

\noindent where $\beta_a$ and $\beta_m$ are used to weight the loss from the acoustic and multimodal views, respectively. These weights can vary along the epochs to facilitate the optimization of the acoustic view. We discuss this in Section \ref{sec:multiview_experiment}, and the training procedure is described in Algorithm \ref{alg:training_algorithm}.


\begin{algorithm}
    \caption{Multi-view Training Algorithm}
    \label{alg:training_algorithm}
    \begin{algorithmic}[1]
        \small
        \Procedure{GetNegSamples}{$Data, \mathrm{y}$}
            \LineComment{Loop through the targets of the batch} 
            \For{$i \gets 1, \dots, \|\mathrm{y}\|$}
                \LineComment{Randomly pick sample with class other than $\mathrm{y}_i$} 
                \State $y^-_i \gets$ \Call{Rand}{$Data$}  
                \Statex[4] ~s.t.~ $y^-_i \neq \mathrm{y}_i$ and 
                \Statex[5] ~~$y^-_i, \mathrm{y}_i$ are acoustically similar
                \LineComment{Collect the corresponding negative inputs} 
                \State $(\mathrm{x}^-_{a_i}, \mathrm{x}^-_{l_i}) \gets getinput(y^-_i)$
            \EndFor
            \State \textbf{return} $(\mathrm{x}^-_{a}, \mathrm{x}^-_{l})$
        \EndProcedure
        
        \Repeat{:}
            \LineComment{Loop through the training batches} 
            \For{$(\mathrm{x}_a, \mathrm{x}_l, \mathrm{y}) \gets nextbatch(Data)$}
                \LineComment{Get the negative acoustic and lexical inputs} 
                \State $(\mathrm{x}^-_a, \mathrm{x}^-_l) \gets$ \Call{GetNegSamples}{$Data, \mathrm{y}$}
                \LineComment{Get the neg. hidden vectors from neg. inputs} 
                \State $\mathrm{h}^-_{a} \gets hidden(V_a, \mathrm{x}^-_a)$ 
                \State $\mathrm{h}^-_{m} \gets hidden(V_m, \mathrm{x}^-_a, \mathrm{x}^-_l)$ 
                \LineComment{Get the pos. hidden vectors and predictions} 
                \State $(\mathrm{h}_{a}, \mathrm{\hat{y}}_{a}) \gets forward(V_a, \mathrm{x}_a)$ 
                \State $(\mathrm{h}_{m}, \mathrm{\hat{y}}_{m}) \gets forward(V_m, \mathrm{x}_a, \mathrm{x}_l)$
                \LineComment{Calculate and add the individual losses} 
                \State $\mathcal{L}_{c} \gets$ \Call{Contrastive}{$\mathrm{h}_{a}, \mathrm{h}^-_{a}, \mathrm{h}_{m}, \mathrm{h}^-_{m}$}
                \State $\mathcal{L}_{a} \gets$ \Call{CrossEntropy}{$\mathrm{y}, \mathrm{\hat{y}}_{a}$}
                \State $\mathcal{L}_{m} \gets$ \Call{CrossEntropy}{$\mathrm{y}, \mathrm{\hat{y}}_{m}$}
                \State $\mathcal{L} \gets \mathcal{L}_{c} + \beta_a \mathcal{L}_{a} + \beta_m \mathcal{L}_{m}$
                \LineComment{Update the parameters using backprop.} 
                \State $\Theta_{V_m} \gets \Theta_{V_m} - \alpha \partial \mathcal{L} / \partial \Theta_{V_m}$
                \State $\Theta_{V_a} \gets \Theta_{V_a} - \alpha \partial \mathcal{L} / \partial \Theta_{V_a}$
            \EndFor
        \Until{stopping criteria met}
    \end{algorithmic}
\end{algorithm}

\noindent \textbf{Teacher-student learning}. 
We anticipate two potential problems with the previously described setting:
1) the learning process may predominantly concentrate on the multimodal view because it has more learning capabilities (i.e., large number of parameters) than the acoustic view, leaving the acoustic model to be of secondary importance during training, 
and 
2) a cross-entropy loss over one-hot vectors ignores informative overlaps among the emotion classes resulting in a very strict objective function. To address these issues, we look into a teacher-student learning approach \citep{softlabelsbetterthanhard}. Given an already-optimized multimodal model $V_m$ (the teacher), we want our acoustic view $V_a$ (the student) to predict probability distributions such as the ones generated by the teacher. We can calculate the difference between the probability distributions of the teacher and the student using Kullback-Leibler (KL) divergence. Then, we minimize the following loss function:
\begin{align} \label{eq:kl_div_loss}
	\mathcal{L}_{KL} = - \frac{1}{N}\sum_i^N p(y_i|x_{m_i}, V_m) log\frac{p(y_i|x_{m_i}, V_m)}{p(y_i|x_{a_i}, V_a)}
\end{align}
where $x_{m_i}$ and $x_{a_i}$ are the multimodal and acoustic inputs for sample $i$, respectively, and $V_m$ and $V_a$ represent the parameters of the views.

\section{Experiments}

We describe the dataset used for the experiments in Section \ref{sec:data_and_feats}. Then, we define the experimental models in Section \ref{sec:experimental_models}, which are used in the multimodal and multi-view experiment in Sections \ref{sec:multimodal_experiment} and \ref{sec:multiview_experiment}.


\subsection{Dataset}
\label{sec:data_and_feats}

\begin{table}[t!]
	\centering
	\renewcommand{\arraystretch}{0.7}
	\setlength{\tabcolsep}{5pt}
    \small
	\begin{tabular}{rllll}
	\toprule
		\textbf{Utterances}	& \textbf{Anger}	& \textbf{Happiness}	& \textbf{Neutral}	& \textbf{Sadness} \\\midrule
        F1 - 528~~	  & 147	& 132	& 171	& 78  \\                        
        M1 - 556~~	  & 82	& 146	& 212	& 116 \\\midrule
        F2 - 479~~	  & 67	& 166	& 134	& 112 \\                        
        M2 - 542~~	  & 70	& 161	& 227	& 84  \\\midrule
        F3 - 522~~	  & 92	& 128	& 130	& 172 \\                        
        M3 - 624~~	  & 148	& 154	& 190	& 132 \\\midrule
        F4 - 527~~	  & 205	& 185	& 75	& 62  \\                        
        M4 - 501~~	  & 122	& 118	& 180	& 81  \\\midrule
        F5 - 590~~	  & 78	& 159	& 221	& 132 \\                        
        M5 - 651~~	  & 92	& 283	& 163	& 113 \\\midrule
        5,520~~   & 1,103	& 1,632	& 1,703	& 1,082 \\
    \bottomrule
	\end{tabular} 
    \caption{Data distribution of the USC-IEMOCAP dataset. F and M mean female and male speakers followed by their session number. }
	\label{tab:eimocap_distr}
\end{table}
\begin{table*}[t!]
\centering
	\renewcommand{\arraystretch}{0.9}
\setlength{\tabcolsep}{10pt}
\begin{tabular}{llllll}
\toprule
\textbf{Type}	& \textbf{Experiment}	& \textbf{Modality} & \textbf{Dev}	& \textbf{Test}	& \textbf{Comment} \\
\midrule
\multirow{5}{*}{Baseline}		& B-ACO-1		& \multirow{2}{*}{Acoustics}	& 0.5858	& -     	& Silence frames \\
                                & B-ACO-2		& ~								& 0.5729	& -     	& Silence frames removed \\ 
\cmidrule{2-6}
                                & B-LEX		    & Lexical						& 0.6706	& -     	& - \\
\cmidrule{2-6}
                                & B-MM-1		& \multirow{2}{*}{Multimodal}	& 0.7195	& -	        & Silence frames \\ 
                                & B-MM-2		& ~								& 0.7265	& -      	& Silence frames removed \\ 
\midrule
\multirow{5}{*}{Hierarchical}	& H-ACO-1		& Acoustics						& 0.5697	& -     	& Acoustic words \\
\cmidrule{2-6}
                                & H-MM-1		& \multirow{4}{*}{Multimodal}	& 0.7316	& -	        & Aligned words  \\
                                & H-MM-2		& ~								& 0.7341	& -     	& ~~~+ GMU \\
                                & H-MM-3		& ~								& 0.7354	& -     	& ~~~+ Attention \\
                                & H-MM-4		& ~								& \bf0.7383	& \bf0.7169	& ~~~+ GMU + Attention \\
\midrule
SOTA							& -      		& Multimodal					& -			& \textbf{0.7079}	& \citet{DBLP:journals/corr/abs-1803-07427} \\
\bottomrule
\end{tabular}
\caption{The results of the multimodal experiments. The name of the experiments starts either with B or H referring to baseline or hierarchical models. ACO, LEX, and MM mean acoustic, lexical and multimodal. Our results provide a new state-of-the-art UA when we use the hierarchical model with GMU and attention. Once the models are optimized on the validation set, we evaluate the best ones on the test set.}
\label{tab:multimodal_results}
\end{table*}

We focus our experiments on the USC-EIMOCAP dataset \citep{busso2008iemocap}. This dataset provides conversations between female and male speakers throughout five sessions. Each session involves a different pair of speakers, which accounts for a total of 10 speakers. The conversations are split into small utterances that map to emotion categories. The original emotion categories are merged to mitigate the unbalanced classes into four categories: \textit{anger}, \textit{happiness}, \textit{neutral}, and \textit{sadness}. Table \ref{tab:eimocap_distr} shows the distribution of the dataset. We split the dataset using the one-speaker-out experimental setting. That is, we take four sessions for training, and the remaining session is split by speakers into the validation and test sets. We report our unweighted accuracy scores running 10-fold cross-validation experiments and averaging scores across folds.


\subsection{Defining Experimental Models }
\label{sec:experimental_models}

\noindent \textbf{B-ACO}: The acoustic baseline is composed of two BLSTM layers of 256 dimensions each, followed by average pooling and a softmax layer. B-ACO-1 uses the raw sequence of frames, whereas B-ACO-2 employs the frames that correspond to the speaker. 

\noindent \textbf{B-LEX}: The lexical baseline uses word embeddings of 1,024 dimensions from ELMo. We feed these vectors into two BLSTM layers of 256 dimensions followed by average pooling and a softmax layer.

\noindent \textbf{B-MM}: The multimodal baseline uses BLSTMs with average pooling over time on each modality, similar to B-ACO and B-LEX. We concatenate the vectors from each modality and feed them into a softmax layer.

\noindent \textbf{H-ACO}: The hierarchical acoustic model uses acoustic word representations. The acoustic words are generated with two BLSTMs of 256 dimensions using the speaker frames (i.e., no silence). At the word level, we perform average pooling over time and feed the resulting vector into a softmax layer.

\noindent \textbf{H-MM}: The hierarchical multimodal model uses the acoustic word representations in H-ACO, and the lexical word representations in B-LEX, with 256 dimensions each. H-MM-1 uses two layers of BLSTM over the concatenated word representations followed by average pooling and a softmax layer. Based on H-MM-1, H-MM-2 incorporates the GMU unit and H-MM-3 adds the attention layer. H-MM-4 uses both GMU and the attention layer.

\subsection{Multimodal Experiments }
\label{sec:multimodal_experiment}

\begin{figure*}[t!]
	\centering
	\includegraphics[width=\linewidth]{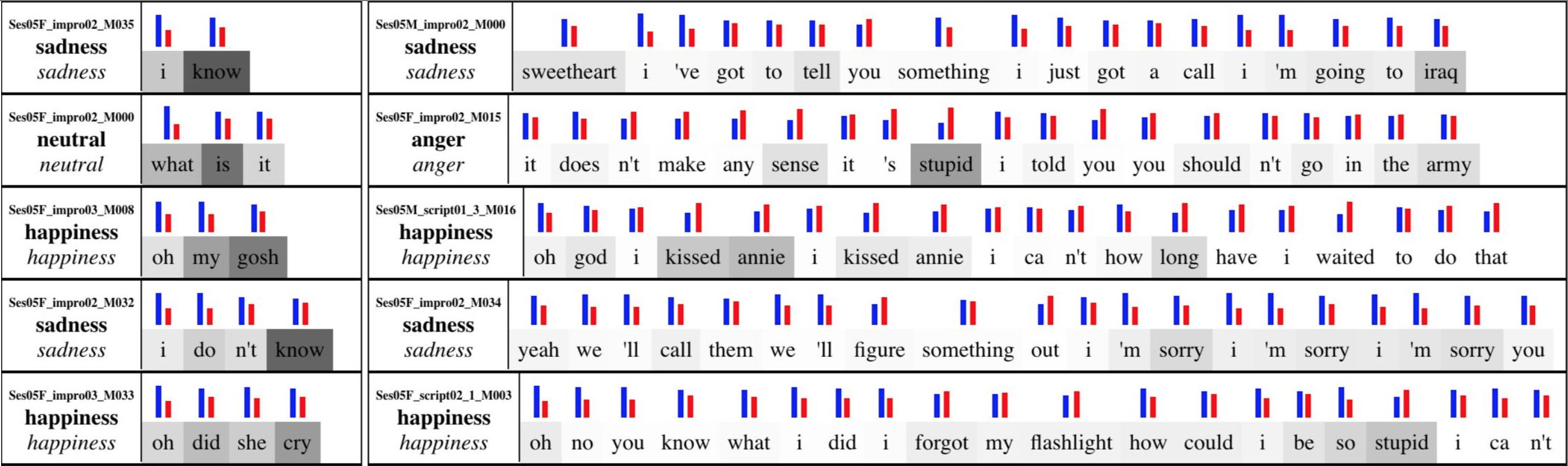}
	\caption{ Multimodal Attention. The figure shows the attention mechanisms at the modality and utterance levels. The bars over the words are the average of $z$ in Eq. \ref{eq:attention-modality}, and they show how much acoustic (left bar in blue) or lexical (right bar in red) information was used. The highlights in the background of the words are the attention probabilities, where the higher the probability the darker the word. }
	\label{fig:multimodal_global_attention}
\end{figure*}

\noindent\textbf{Impact of silence}. 
We experiment with silence and the baselines B-ACO and B-MM. In Table \ref{tab:multimodal_results}, although keeping silence seems better than removing it (B-ACO-1 vs. B-ACO-2), the multimodal model shows a small improvement when silence is ignored (B-MM-1 vs. B-MM-2). 
By looking into the predictions, besides the silence and environmental noise in the original frames, we notice that a second speaker can influence the emotions of the speaker being evaluated. This observation, along with the model improvements, suggests that is possible to fuse information more efficiently.

\noindent\textbf{Hierarchical models}. To make better use of the modalities, we align lexical information with acoustic representations at the word level. Based on the silence impact, our acoustic word representations only use frames where the speaker intervenes in the conversation (i.e., no silence or other speakers). Similar to the previous scenario, we see a detrimental behavior in the hierarchical acoustic model compared to the models that use the original sequence of frames (H-ACO-1 vs. B-ACO). However, when we concatenate the lexical and acoustic word representations (H-MM-1), our hierarchical model surpasses the UA of all previous models. In fact, our best model (H-MM-4) outperforms the previous state-of-the-art UA. This serves as strong evidence that fusing information more efficiently can yield a better performance.

\noindent\textbf{Ablation experiment}. Table \ref{tab:multimodal_results} shows the performance of the hierarchical multimodal models with and without the modality- and context-based attention mechanisms (H-MM). Using H-MM-1 as a common ground, the modality-based attention (H-MM-2) provides an improvement of about 1\% on the UA metric. This result suggests that one modality can be more informative than the other, and hence, it is important to prioritize the one that carries more emotional information. Likewise, adding the attention mechanism, H-MM-3, outperforms H-MM-1 by a similar percentage. Our intuition is that weighting the words that provide strong emotional information based on the context allows the model to disambiguate meaning and discriminate more easily the samples. Lastly, H-MM-4 combines both attention mechanisms, which improves over the individual attention models H-MM-2 and H-MM-3 by about 1\% of UA. This means that the attention mechanisms are more complementary than overlapping. 
 
\noindent\textbf{Attention visualization}. 
For the modality-based attention, the vector $z$ from Eq. \ref{eq:attention-modality} determines how much acoustic information will go through the next layers, whereas $(1-z)$ is the amount of lexical data allowed. 
Figure \ref{fig:multimodal_global_attention} provides a visualization of these vectors. The bars show the amount of information that is captured from one modality versus the other. For instance, the sample \textit{``oh my gosh''} illustrates that the words rely on more acoustic than lexical information. Intuitively, this phrase by itself could describe different emotions, but it is the acoustic modality that mitigates the ambiguity. 
Regarding the context-based attention, Figure \ref{fig:multimodal_global_attention} shows the places where the model focuses along the utterance. For large-context utterances, where the acoustic features are more or less similar, the semantics can help to highlight specific spots.  
For example, in the second sentence on the right of Figure \ref{fig:multimodal_global_attention}, the model detects the semantics of the words \textit{sense} and \textit{stupid} and associates them with the words \textit{should}, \textit{go}, and \textit{army}. The attention mechanism not only emphasizes semantics but it also takes into account the acoustic features. In the same block of sentences, it is worth noting that the words primarily driven by acoustics (e.g., \textit{sweatheart}, \textit{oh god}, \textit{sorry} and \textit{yeah}) are highlighted by the attention mechanism. These results also align with the intuition that the attention mechanisms are complementary.

\begin{table*}[t!]
\centering
\renewcommand{\arraystretch}{1.1}
\setlength{\tabcolsep}{9pt}
\begin{tabular}{lllll}
    \toprule
    \bf View1 & \bf View 2 & \bf Dev & \bf Test & \bf Comment\\
    \midrule
    B-ACO-1 & - & 0.5858	& 0.5443 & Acoustic-exclusive baseline \\
    \midrule
    \multirow{4}{*}{B-ACO-1} 
        & B-LEX
        & 0.5971 
        & - 
        & Loss: $\mathcal{L}_c + \mathcal{L}_a + \mathcal{L}_m$ 
        (Eqs. \ref{eq:contrastive_loss}, \ref{eq:aco-cross-entropy}, \ref{eq:lex-cross-entropy}) \\
        & H-MM-4
            & 0.5976        
            & - 
            & Loss: $\mathcal{L}_c + \mathcal{L}_a + \mathcal{L}_m$ 
            (Eqs. \ref{eq:contrastive_loss}, \ref{eq:aco-cross-entropy}, \ref{eq:lex-cross-entropy}) \\
        & H-MM-4 $\dagger$
            & 0.5969      
            & - 
            & Loss: $\mathcal{L}_c + \mathcal{L}_a$ 
            (Eqs. \ref{eq:contrastive_loss}, \ref{eq:aco-cross-entropy}) \\
        & H-MM-4 $\dagger$
            & \bf0.6060   
            & \bf0.5859 
            & Loss: $\mathcal{L}_c + \mathcal{L}_{KL}$ (Eqs. \ref{eq:contrastive_loss}, \ref{eq:kl_div_loss}) \\
    \midrule                    
    B-ACO-1 + Attention 
        & H-MM-4 $\dagger$
            & \bf0.6100 
            & \bf0.5976 
            & Loss: $\mathcal{L}_c + \mathcal{L}_{KL}$  (Eqs. \ref{eq:contrastive_loss}, \ref{eq:kl_div_loss}) \\
    \bottomrule
\end{tabular}
\caption{The results of the multi-view experiments. We use the acoustic model B-ACO-1 as the first view and evaluate its performance using different second views. $\dagger$ means that the second view is not updated during training and its classification loss is not included.
}
\label{tab:multiview_results}
\end{table*}



\subsection{Multi-view experiments}
\label{sec:multiview_experiment}

Our multi-view experiments use utterance-level representations to calculate the contrastive loss in Eq. \ref{eq:contrastive_loss}. We discard experiments at the word level because 1) contrasting emotions for every word individually poses a complex task\footnote{Negative words are hard to choose because we want properly formed utterances with the same number of words.}, and 2) context helps to disambiguate meaning as well as to convey the overall emotion rather than relying on high emotional words individually. Additionally, our experiments aim at a more practical scenario where there is no need for transcripts or ASR output with forced alignment. 


\noindent\textbf{Choosing negative samples}. 
To calculate the loss as in Eq. \ref{eq:contrastive_loss}, we randomly choose negative samples in two ways: 1) forcing a different class, and 2) forcing a different class that is acoustically similar to the positive sample (e.g., \textit{sadness} vs. \textit{neutral}, or \textit{anger} vs. \textit{happiness}). 
We saw that the model generalizes better using the second option. Our intuition is that the model does not have problems to force the margin $m$ between vectors when the negative input samples come from fairly easy discriminative classes (e.g., \textit{happiness} vs. \textit{neutral}). In contrast, the model struggles to force the margin $m$ between vectors when classes are acoustically similar, which turns into better generalization. 

\noindent\textbf{Different views}. We choose B-ACO-1 as the first view because it uses raw frame level features. As shown in Table \ref{tab:multiview_results}, we compare B-LEX and H-MM-4 as simple and elaborated second views by applying the contrastive and the views' cross-entropy loss functions. Indeed, by using B-LEX we show that the acoustic model B-ACO-1 improves its accuracy. Further improvements are made if we use H-MM-4 as a second view. This means that it is better to transfer information to the acoustic model when the modalities are effectively combined rather than when we try to induce only lexical information.

\noindent\textbf{Frozen weights}. We further explore H-MM-4 as a second view by first optimizing it, and then fixing its weights in the multi-view setting. 
Experiments with a trainable second view show that the lexical model is prioritized even when the losses are weighted as in Eq. \ref{eq:aco-cross-entropy} and \ref{eq:lex-cross-entropy}. The intuition is that there is nothing new that this second view can learn from the multi-view setting once it has been optimized separately, and thus, it is better to exclude the complexity of learning it from scratch. Table \ref{tab:multiview_results} shows a small improvement over the previous models reaching 59.69\% of UA on the validation set.

\noindent\textbf{Teacher-student learning}. We also experiment with a teacher-student setting where the model H-MM-4 is optimized separately. This model is a non-trainable second view where its class predictions are used as soft labels to evaluate the first view. The idea is to provide informative similitudes among the training samples by evaluating against a probability distribution over the classes rather than hard labels. The model reduces its loss more steadily than previous models, and once optimized, it surpasses previous results. Finally, we consider the case of a more complex student network since previous studies suggest that small student models may not be able to cope with the teacher models \citep{softlabelsbetterthanhard, adversarialteacherstudent}. By adding an attention layer over the acoustic model B-ACO-1, we are able to improve the accuracy of the model by 1\% absolute points, as shown in Table \ref{tab:multiview_results}.



\section{Conclusions}

We presented multimodal and multi-view approaches for emotion recognition. The first approach assumes that lexical information is always available when the speech signal is being processed. For such a scenario, our hierarchical multimodal model outperforms the state-of-the-art score with the aid of modality- and context-based attention mechanisms. The second approach adapts to a more realistic scenario where lexical data may not be available for evaluation. Our multi-view setting has shown that acoustic models can still benefit from lexical information over models that have been exclusively trained on acoustic features.

\bibliography{acl2019}
\bibliographystyle{acl_natbib}

\appendix
\section*{\centering Appendix for ``Multimodal and Multi-view Models for Emotion Recognition''}

\section{Dataset Insights}

This section describes some insights of the dataset. We use this information to take decisions relevant to our experiments. We consider the maximum number of words, the length of frames per utterance, and the number of frames per words.

\begin{figure}[H]
	\centering
	\includegraphics[width=\linewidth]{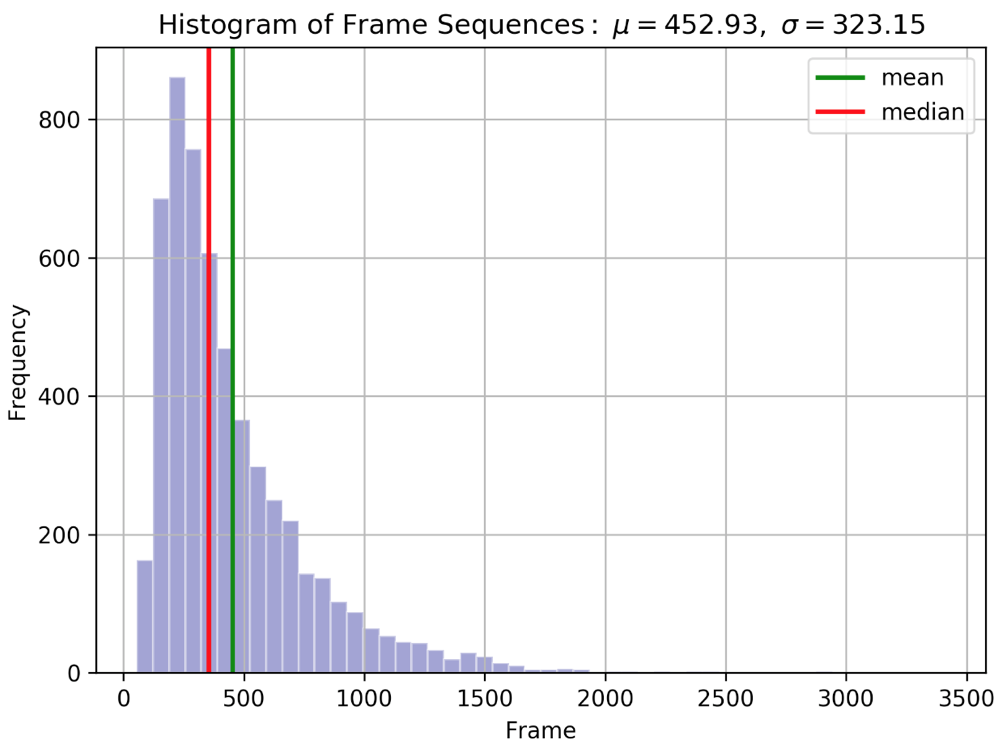}
	\caption{Histogram of frame sequences for every utterance.}
	\label{app:fig:hist_frame_seq}
\end{figure}

\begin{figure}[H]
	\centering
	\includegraphics[width=\linewidth]{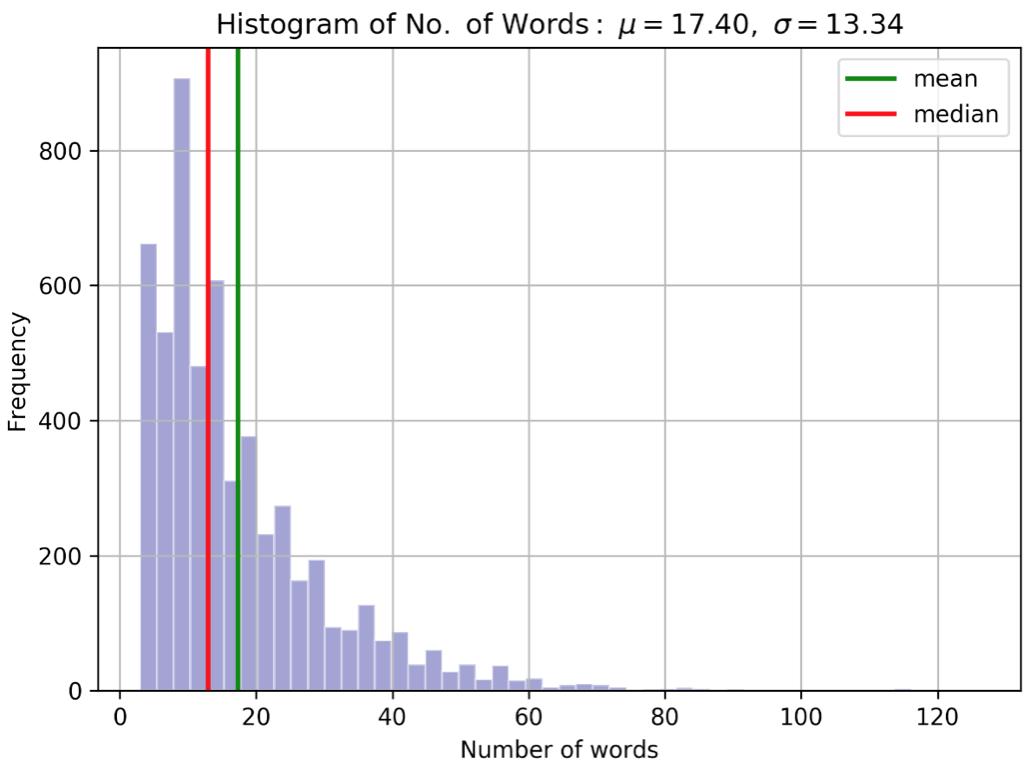}
	\caption{Histogram of number of words per utterance.}
	\label{app:fig:hist_no_words}
\end{figure}

We use 30 words as a maximum length for the sentences given that he average length is 17.40 and the standard deviation of 13.34 (see Figure \ref{app:fig:hist_no_words}). Additionally, we show statistics for the frame lengths on each utterance in Figure \ref{app:fig:hist_frame_seq}. We take a maximum length of 700 frames per utterance, where each frame is equivalent to 10 milliseconds. 

\begin{figure}[H]
	\centering
	\includegraphics[width=\linewidth]{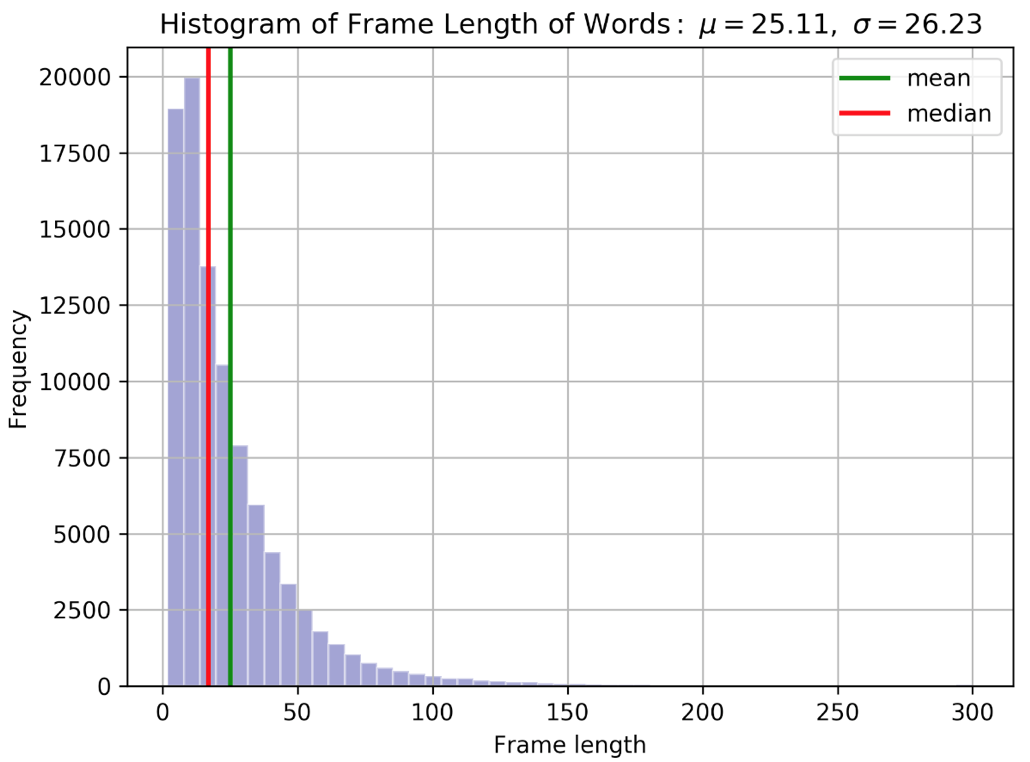}
	\caption{Histogram of word lengths in terms of frames.}
	\label{app:fig:hist_frame_length_words}
\end{figure}

We also obtain the average length of frames that each word has according to the alignments of the dataset. Note that most of the words are within 100 frames, or equivalently, 1 second (see Figure \ref{app:fig:hist_frame_length_words}).

\section{Experimental Settings}
\label{app:sec:experimental_settings}

We train all our models for 30 epochs using a learning rate of 1e-4 and a batch size of 64. The optimization of the models is conducted using Adam \citep{DBLP:journals/corr/KingmaB14}. We consistently use gradient clipping among our experiments. We clip the norm of the gradient beyond 5 \citep{DBLP:journals/corr/abs-1211-5063, Goodfellow-et-al-2016}:
$$\mathbf{g} \leftarrow \frac{\mathbf{g}\tau}{||\mathbf{g}||} ~~\mathrm{if}~ ||\mathbf{g}|| > \tau$$

To regularize the models, we use dropouts \citep{Srivastava:2014:DSW:2627435.2670313} by choosing drop probabilities between 0.4 and 0.5. We apply an $\ell_2$ with a coefficient of 1e-5. For the GMU component, we use batch normalization applied to each modality matrix \citep{DBLP:journals/corr/IoffeS15}. All our experiments are validated using 10-fold cross-validation, leaving one speaker out of the training and validation sets. 

For the multi-view learning experiments, we use the same settings as described for the multimodal experiments. In the case of the loss weights $\beta_a$ and $\beta_m$, we experiment with values in \{1.0, 1.2\} and \{0.3, 0.5, 1.0\}, respectively. We also experiment with $\beta$s as function of the epochs using 
$$\beta = \frac{1}{1 + (\rho * epoch)} \beta_{o}$$ 

\noindent where $\rho$ is a decreasing rate and $\beta_{o}$ is the initial value, but the learning setting still overemphasize the multimodal view. The best results were achieved with $\beta_a=1$ and $\beta_m=0.3$ when both views were optimized simultaneously. For the margin in the contrastive loss function, we use $m=0.5$.

For negative sampling in the contrastive loss function, we empirically found that using \texttt{anger} with \texttt{happiness} and \texttt{neutral} with \texttt{sadness} generally worked well since the acoustic patterns are similar. However, we saw some informative pairs when \texttt{happiness} and \texttt{anger} were coupled with \texttt{neutral}. This suggests that a more systematic way to determine pairs is needed. We leave the exploration of metrics such as valence, arousal and dominance to determine the contrastive pairs for future work.

\section{Additional Experiments}
\label{app:sec:side_experiments}

We run the following side experiments:

\begin{itemize}
    \item Different length of words for our lexical baseline model (B-LEX). No benefit was perceived by going beyond 30 words.
    \item Different length of frames for our acoustic baseline model (B-ACO). The training time increases significantly while there is no substantial gain on performance by doing this.
    \item Improvised versus scripted utterances. We saw a substantial increase in performance (~3\%) of UA when speakers use scripted language rather than natural conversations.
\end{itemize}

\section{Model Insights}

\subsection{Visualization of Attention}

\begin{figure*}[t!]
	\centering
	\includegraphics[width=\linewidth]{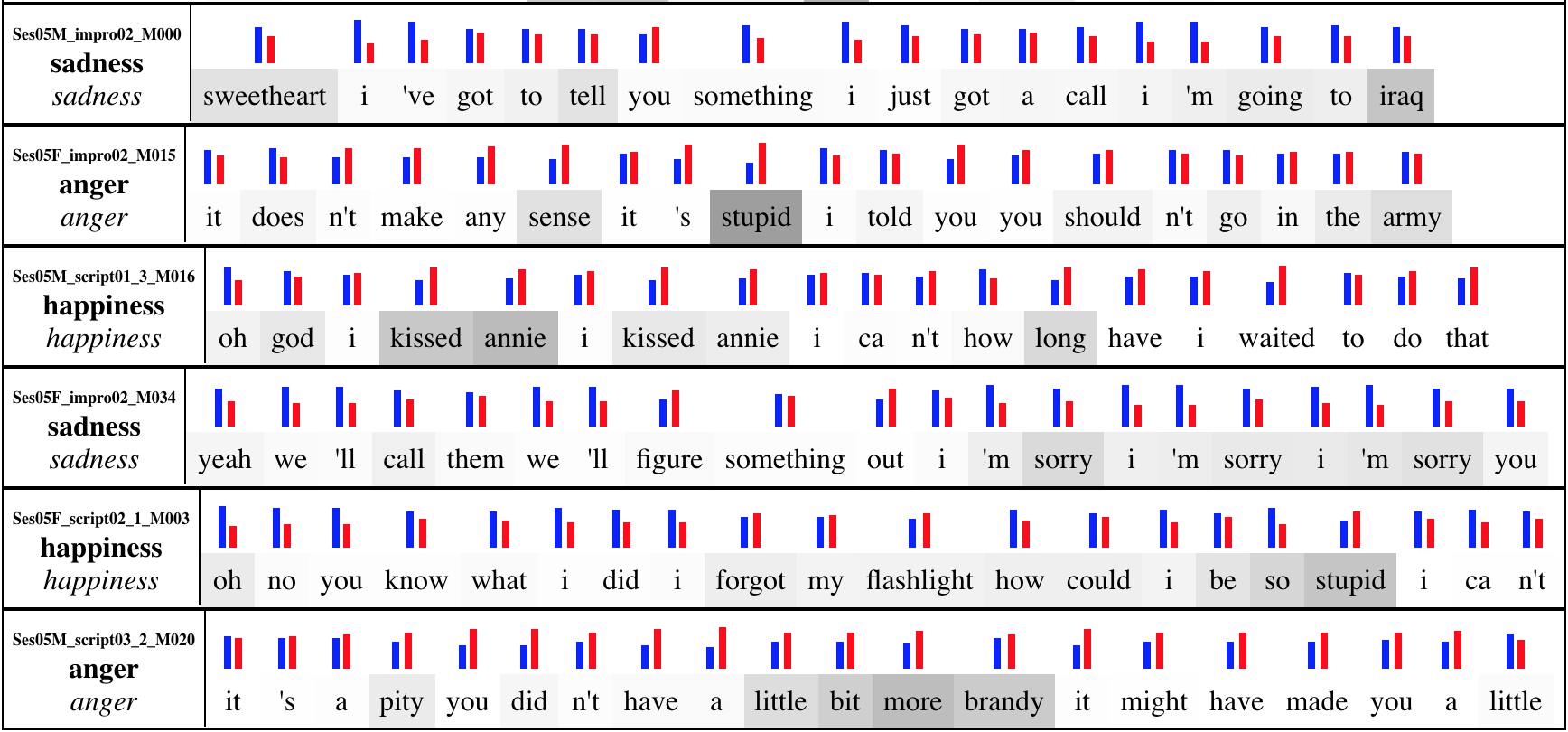}
	\caption{Correct predictions (italics) of the model along with the attention visualization.}
	\label{app:fig:attention_viz_correct}
\end{figure*}

\begin{figure*}[t!]
	\centering
	\includegraphics[width=\linewidth]{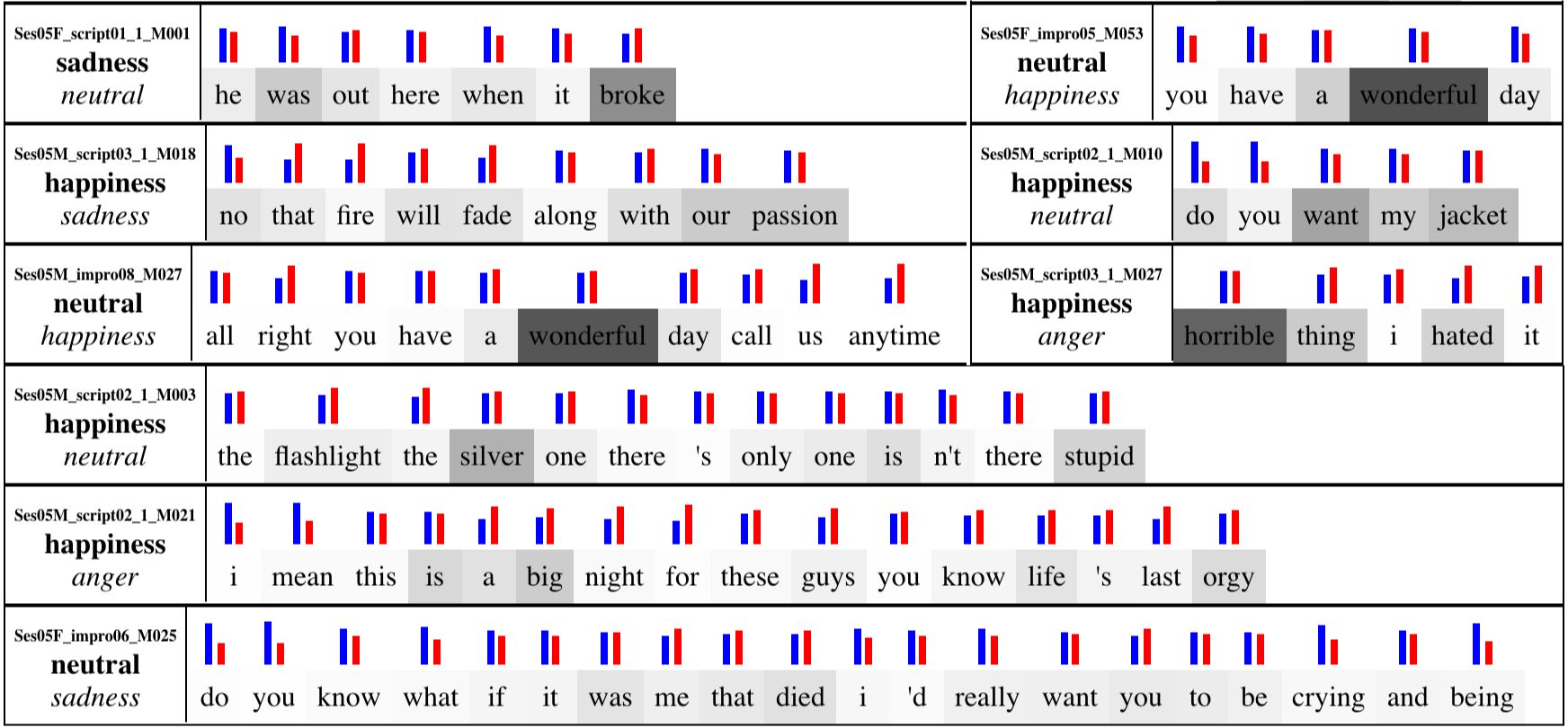}
	\caption{Incorrect predictions (italics) of the model along with the attention visualization.}
	\label{app:fig:attention_viz2_incorrect}
\end{figure*}

We visualize the attention weights for correctly and incorrectly predicted emotions in Figures \ref{app:fig:attention_viz_correct} and \ref{app:fig:attention_viz2_incorrect}. Interestingly, when the sentences are read by humans, the target emotion for such utterances turn out ambiguous, which aligns with the result of the models.

\subsection{Multi-view Results}

\begin{figure}[t!]
	\centering
	\includegraphics[width=\linewidth]{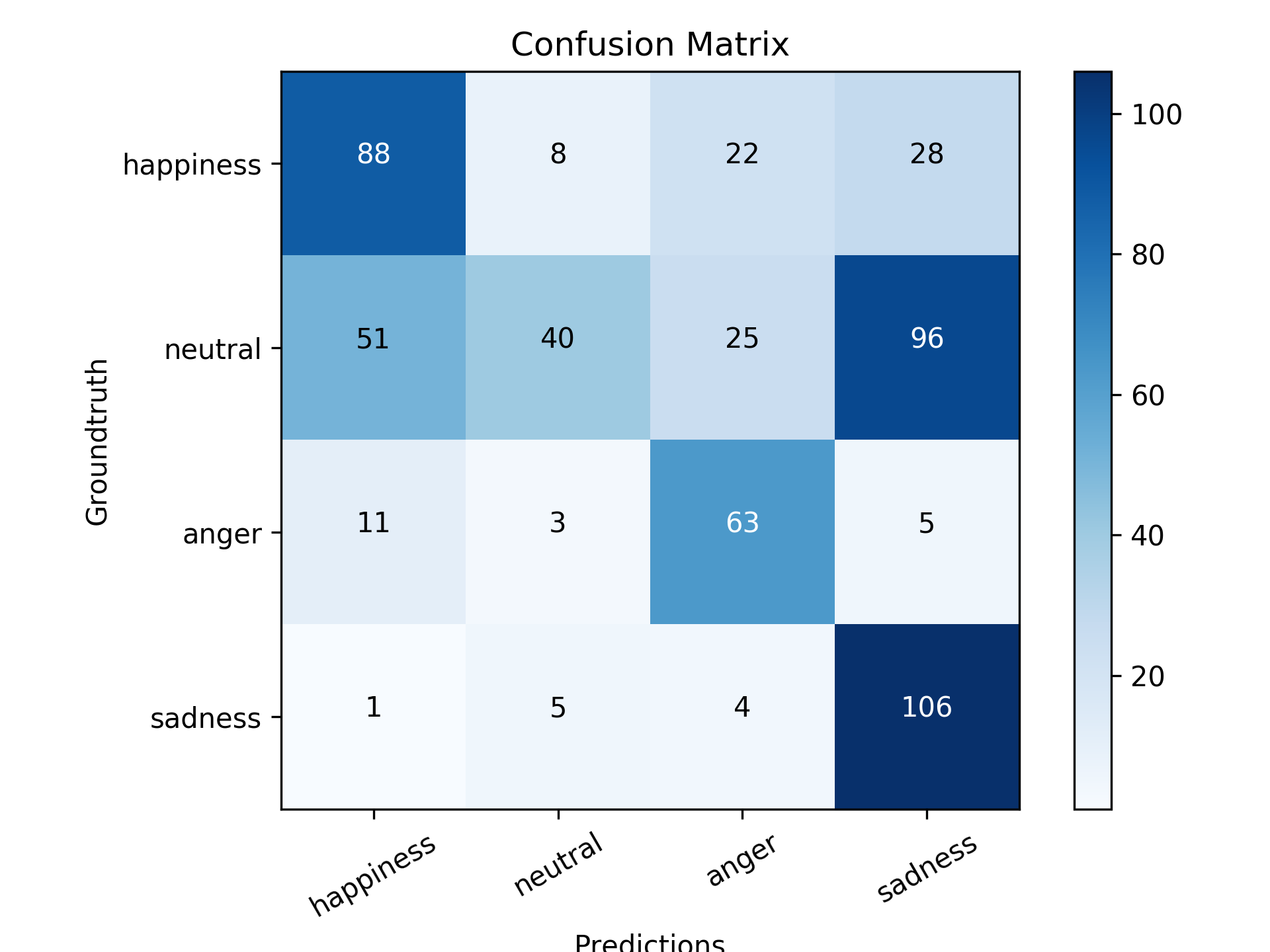}
	\caption{Confusion matrix of the acoustic model B-ACO-1.}
	\label{app:fig:aco_conf_mat}
\end{figure}

\begin{figure}[t!]
	\centering
	\includegraphics[width=\linewidth]{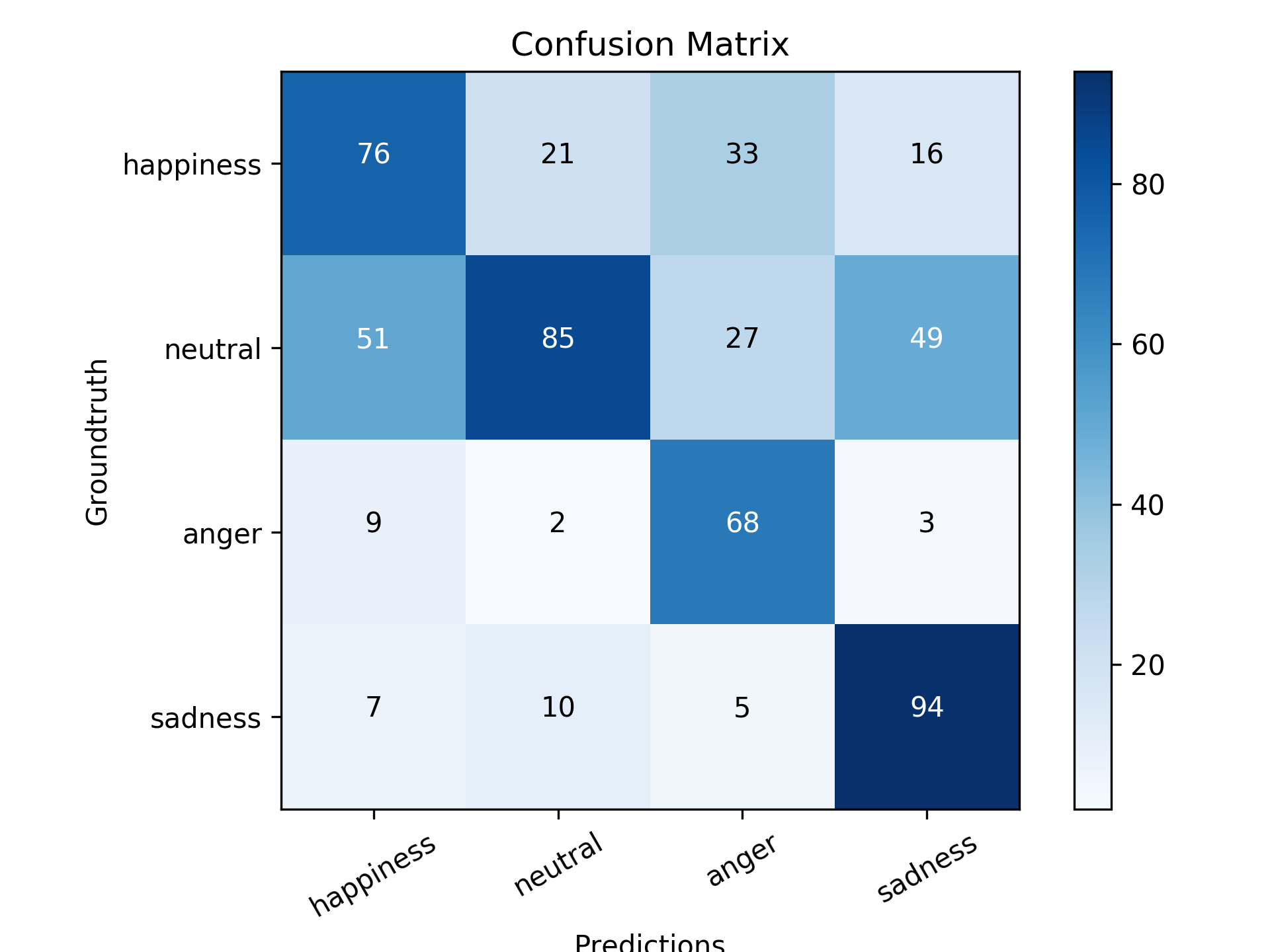}
	\caption{Confusion matrix of the acoustic model B-ACO-1 trained in a multi-view learning setting.}
	\label{app:fig:mv_conf_mat}
\end{figure}

By using the multi-view learning setting, we manage to induce lexical information into the model. According to Figures \ref{app:fig:aco_conf_mat} and \ref{app:fig:mv_conf_mat}, it is easy to see that the model B-ACO-1 corrects a lot of the mistakenly predicted classes (i.e., compare neutral as ground-truth and sadness as prediction). However, the images also reveal that there are side effects such as transferring wrong aspects of the lexical modal to the acoustic one.
\end{document}